\documentclass[11pt,a4paper]{article}
\usepackage{times,latexsym}
\usepackage{url}
\usepackage[T1]{fontenc}

\usepackage{algorithm}
\usepackage{algpseudocode}
\usepackage{algorithmicx}
\usepackage{varwidth}
\usepackage{float}
\usepackage{graphicx}
\usepackage{subfig}
\usepackage{multirow}
\usepackage{amsmath}
\usepackage{amssymb}
\usepackage{amsthm}
\usepackage{makecell}
\usepackage{tabularx}
\newtheorem{theorem}{Theorem}[section]
\usepackage[acceptedWithA]{tacl2018v2}

\usepackage{xspace,mfirstuc,tabulary}

\newif\iftaclinstructions
\taclinstructionsfalse 
\iftaclinstructions

\renewcommand{\anonsubtext}{(No author info supplied here, for consistency with
TACL-submission anonymization requirements)}
\newcommand{\instr}
\fi

\iftaclpubformat 

\else

\fi

\newcolumntype{Y}{>{\arraybackslash}X}

\title{Minimum Description Length Recurrent Neural Networks}

\author{
Nur Lan\textsuperscript{1,2}, Michal Geyer\textsuperscript{2}, Emmanuel Chemla\textsuperscript{1,3,*}, Roni Katzir\textsuperscript{2,*}
\\\textsuperscript{1}Ecole Normale Supérieure
\\\textsuperscript{2}Tel Aviv University\\\textsuperscript{3}EHESS, PSL University, CNRS 
\\ 
\texttt{\{nlan,chemla\}@ens.fr}\\
\texttt{michalgeyer@mail.tau.ac.il}\\ \texttt{rkatzir@tauex.tau.ac.il}}

\date{}

\algdef{SE}[SUBALG]{Indent}{EndIndent}{}{\algorithmicend\ }%
\algtext*{Indent}
\algtext*{EndIndent}

\algnewcommand{\parState}[1]{\State%
    \parbox[t]{\dimexpr\linewidth-\algmargin}{\strut\hangindent=\algorithmicindent \hangafter=1 #1\strut}}

\begin{document}
\pagenumbering{arabic}
\maketitle
\begin{abstract}

We train neural networks to optimize a Minimum Description Length score, i.e., to balance between the complexity of the network and its accuracy at a task. We show that networks optimizing this objective function master tasks involving memory challenges and go beyond context-free languages. 
These learners master languages such as $a^nb^n$, $a^nb^nc^n$, $a^nb^{2n}$, $a^nb^mc^{n+m}$, and they perform addition. 
Moreover, they often do so with 100\% accuracy. 
The networks are small, and their inner workings are transparent. We thus provide formal proofs that their perfect accuracy holds not only on a given test set, but for any input sequence. To our knowledge, no other connectionist model has been shown to capture the underlying grammars for these languages in full generality.

\end{abstract}

\renewcommand*{\thefootnote}{\fnsymbol{footnote}}
\footnotetext[1]{Both authors contributed equally to this work.}
\renewcommand*{\thefootnote}{\arabic{footnote}}

\section{Introduction}

A successful learning system is one that makes appropriate generalizations. For example, after seeing the sequence 1,0,1,0,1 we might suspect that the next element will be 0. If we then see 0, we might be even more confident that the next input element will be 1. Artificial neural networks have shown impressive results across a wide range of domains, including linguistic data, computer vision, and many more. They excel at generalizing when large training corpora and computing resources are available, but they face serious challenges that become particularly clear when generalizing from small input sequences like the one presented above.

First, they tend to overfit the learning data. To avoid this, they require external measures to control their own tendency for memorization (such as regularization) as well as very large training corpora. Moreover, standard regularization techniques fall short in many cases, as we show below.

Second, even when successful, they tend to produce non-categorical results. That is, they output very high probabilities to target responses, but never 100\%. Adequate, human-like generalization, on the other hand involves having both a probabilistic guess (which neural networks can do) and, at least in some cases, a clear statement of a categorical best guess (which neural networks cannot do). 

Third, these networks are often very big, and it is generally very hard to inspect a given network and determine what it is that it actually knows (though see \citealp{lakretz_emergence_2019} for a recent successful attempt to probe this knowledge in the context of linguistics).

Some of the challenges above arise from the reliance of common connectionist approaches on backpropagation as a training method, which keeps the neural architecture itself constant throughout the search. The chosen architecture must therefore be large enough to capture the given task, and it is natural to overshoot in terms of size. Furthermore, it must allow for differentiable operations to be applied, which prevents certain categorical patterns from even being expressible.

In this paper, we propose to investigate a training method which differs from common approaches in that its goal is to optimize a Minimum Description Length objective function (MDL; \citealp{Rissanen:1978}). 
This amounts to minimizing error as usual, while at the same time trying to minimize the size of the network (a similar pressure to a Bayesian size prior). As a result, the objective function offers a guide to determining the size of the network (a guide that error minimization alone does not provide), which means that the architecture itself can evolve during learning and typically can decrease in size. 
One potential side effect is that optimization cannot be done through backpropagation alone. We here use a genetic algorithm to search through the very large search space of neural networks of varying sizes.

We find that MDL-optimized networks reach adequate generalizations from very small corpora, and they avoid overfitting. The MDL-optimized networks are all small and transparent; in fact, we provide proofs of accuracy that amount to infinite and exhaustive test sets. They can also provide deterministic outputs when relevant (expressing pure 100\% confidence). We illustrate this across a range of tasks involving counters, stacks, and simple functions such as addition.

\section{Previous work}

Our primary concern in this paper is the objective function.
The idea of applying a simplicity criterion to artificial neural networks dates back at least to \citet{hinton_keeping_1993}, who minimize the encoding length of a network's weights alongside its error, and to \citet{ZhangMuhlenbein:1993,ZhangMuhlenbein:1995}, who use a simplicity metric that is essentially the same as the MDL metric that we use in the present work (and describe below). \citet{Schmidhuber:1997} presents an algorithm for discovering networks that optimize a simplicity metric that is closely related to MDL. Simplicity criteria have been used in a range of works on neural networks, including recent contributions (e.g., \citealp{AhmadizarSoltanianAkhlaghianTabTsoulos:2015} 
and \citealp{GaierHa:2019}). Outside of neural networks, MDL --- and the closely related Bayesian approach to induction --- have been used in a wide range of models of linguistic phenomena, in which one is often required to generalize from very limited data (see \citealp{Horning:1969}, \citealp{Berwick:1982}, \citealp{Stolcke:1994}, \citealp{Grunwald:1996}, and \citealp{Marcken:1996}, among others, and see \citealp{RasinKatzir:2016} and \citealp{RasinBergerLanShefiKatzir:2021} for recent proposals to learn full phonological grammars using MDL within two different representational frameworks). In the domain of program induction, \citet{yang_one_2022} have recently used a Bayesian learner equipped with a simplicity prior to learn formal languages similar to the ones we present below.

Turning to the optimization algorithm that we use to search for the MDL-optimal network, our work connects with the literature on using evolutionary programming to evolve neural networks. Early work that uses genetic algorithms for various aspects of neural network optimization includes \citet{MillerToddHegde:1989}, \citet{MontanaDavis:1989}, \citet{WhitleyStarkweatherBogart:1990}, and \citet{ZhangMuhlenbein:1993,ZhangMuhlenbein:1995}. These works focus on feed-forward architectures, but \citet{AngelineSaundersPollack:1994} present an evolutionary algorithm that discovers recurrent neural networks and test it on a range of sequential tasks that are very relevant to the goals of the current paper. Evolutionary programming for neural networks remains an active area of research (see \citealp{Schmidhuber:2015} and \citealp{GaierHa:2019}, among others, for relevant references).

Our paper connects also with the literature on using recurrent neural networks for grammar induction and on the interpretation of such networks in terms of symbolic knowledge (often formal-language theoretic objects). These challenges were already taken up by early work on recurrent neural networks (see  \citealp{GilesSunChenLeeChen:1990} and \citealp{Elman:1990}, among others), and they remain the focus of recent work (see, e.g., \citealp{WangZhangOrorbia-IIXingLiuGiles:2018} and \citealp{WeissGoldbergYahav:2018a}). \citet{Jacobsson:2005} and \citet{WangZhangOrorbia-IIXingLiuGiles:2018} provide discussion and further references.

\section{Learner}

\subsection{Objective: Minimum Description Length}

Consider a hypothesis space $\mathcal{G}$ of possible grammars, and a corpus of input data $D$. In our case, $\mathcal{G}$ is the set of all possible network architectures expressible using our representations, and $D$ is a set of input sequences. For a given $G\in \mathcal{G}$ we may consider the ways in which we can encode the data $D$ given $G$. The MDL principle (\citealp{Rissanen:1978}), a computable approximation of Kolmogorov Complexity (\citealp{Solomonoff:1964,Kolmogorov:1965,Chaitin:1966}), aims at the $G$ that minimizes $|G|+|D:G|$, where $|G|$ is the size of $G$ and $|D:G|$ is the length of the shortest encoding of $D$ given $G$ (with both components typically measured in bits). Minimizing $|G|$ favors small, general grammars that often fit the data poorly. Minimizing $|D:G|$ favors large, overly specific grammars that overfit the data. By minimizing the sum, MDL aims at an intermediate level of generalization: reasonably small grammars that fit the data reasonably well. 

The term $|D:G|$ corresponds to the surprisal of the data $D$ according to the probability distribution defined by $G$ (i.e., the negative log of the probability assigned to targets by the network). The term $|G|$ depends on an encoding scheme for mapping networks onto binary strings, described below. 

\subsection{Our networks and their encoding}

The MDL learner explores a space of directed graphs, made of an arbitrary number of units and weighted connections between them. We describe the actual search space explored in the experiments below, by explaining how these networks are uniquely encoded to produce an encoding length $|G|$. 

\subsubsection{Example}

\begin{figure}[H]
\centering
\includegraphics[scale=0.2]{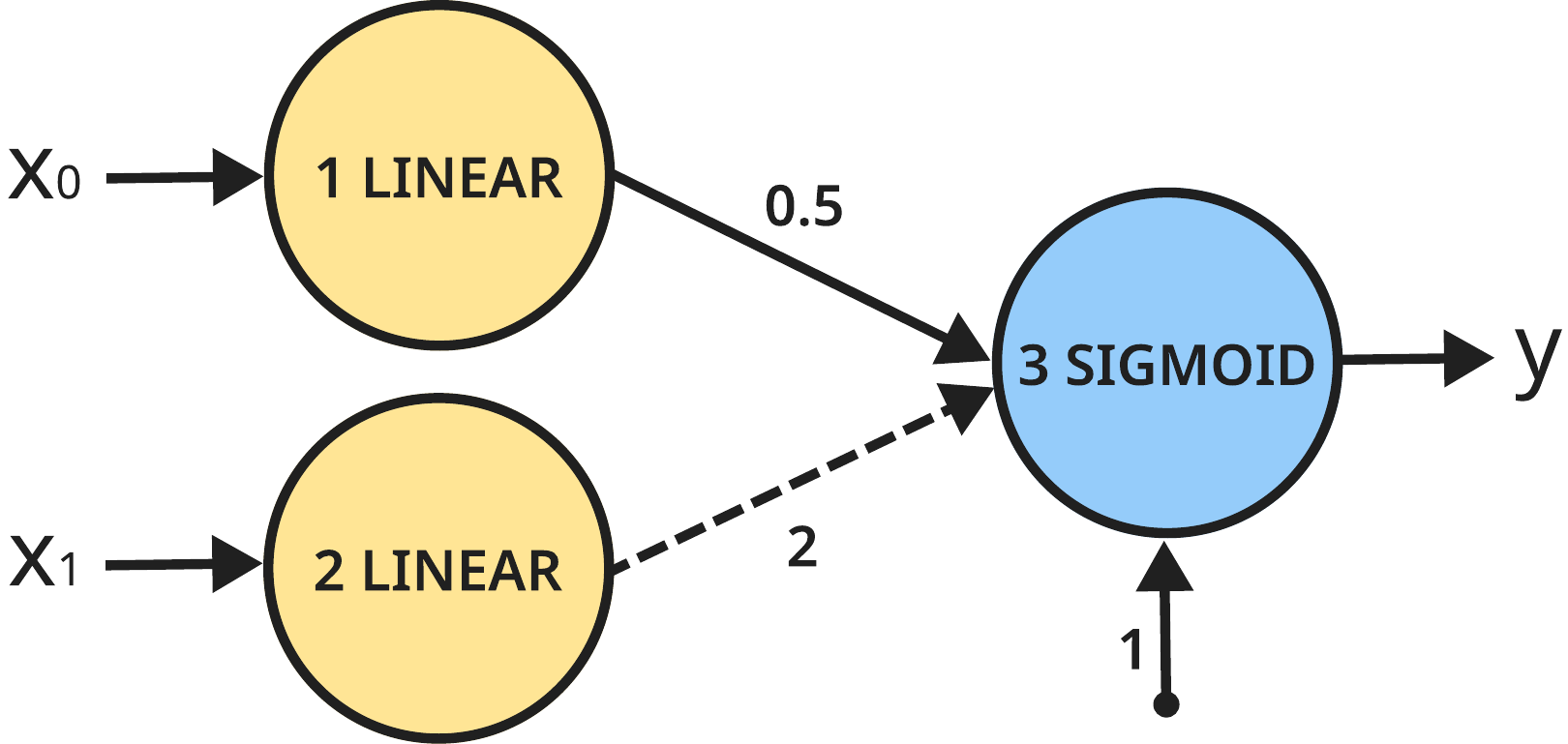}
\caption{Example network, encoded in Fig.~\ref{fig:network_binary_encoding}}
\label{fig:encoded_net}
\end{figure}

We will consider networks such as the one represented in Fig.~\ref{fig:encoded_net}. It consists of two input units (yellow units 1 and 2) and one output unit with a sigmoid activation (blue unit 3). The network has one forward connection (from unit 1 to 3) and one recurrent connection (unit 2 to 3), represented by a dashed arrow. Recurrent connections feed a unit with the value of another unit at the previous time step, and thus allow for the development of memory across the different time steps of the sequential tasks we are interested in. Here, unit 3 is fed with input 2 from the previous step. The connection weights are $w_{1,3} = 0.5$ and $w_{2,3} = 2$. Unit 3 is also fed a bias term $b_3 = 1$ represented by a sourceless arrow.

We will now explain how such a network is represented to measure its encoding size $|G|$.

\begin{figure*}[h]
\centering 
\[
\underbrace{\underbrace{11011}_{E(U)}\underbrace{\underbrace{00}_{\text{linear}}\underbrace{101}_{E(\#out)}\underbrace{00..11}_{\text{unit 3}}\underbrace{01..01}_{w_{1,3}}\underbrace{0}_{\text{forward}}}_{\text{Unit 1}}\underbrace{\underbrace{00}_{\text{linear}}\underbrace{101}_{E(\#out)}\underbrace{00..11}_{\text{unit 3}}\underbrace{01..01}_{w_{2,3}}\underbrace{1}_{\text{recurrent}}}_{\text{Unit 2}}\underbrace{\underbrace{01}_{\text{sigmoid}}\underbrace{0}_{E(\#out)}\underbrace{1111}_{\text{cost}(\text{sigmoid})}\underbrace{11..01}_{bias}}_{\text{Unit 3}}}_{\text{Encoded Network}}
\]
\caption{Binary encoding of the network in Fig.~\ref{fig:encoded_net}}

\label{fig:network_binary_encoding}
\end{figure*}

\subsubsection{Preliminary: encoding numbers}
To ensure unique readability of a network from its string representation we use the prefix-free encoding for integers from \citet{LiVitanyi:2008}:
\[
E(n)=\underbrace{11111...1111}_{\text{Unary enc. of }\lceil{log_2{n}}\rceil\ \ }\underbrace{0}_{\text{Separator}}\underbrace{10101...00110}_{\text{Binary enc. of $n$}}
\]

\subsubsection{Encoding a network}
\label{sec:encoding}
The encoding of a network is the concatenation of (i)~its total number of units, and (ii)~the ordered concatenation of the encoding of each of its units.

\subsubsection{Units}
The encoding of a unit includes
its activation function, the number of its outgoing connections, the encoding of each of its outgoing connections, and its bias weight, if any.

\subsubsection{Activation functions}

Possible activation functions are: the linear activation (identity), ReLU, sigmoid, square, as well as the floor function and the unit step function (returns 0 for inputs $\leq0$ and 1 otherwise). To build an intuitive measure of simplicity into the model’s choice of activation functions, we add a cost to each function, encoded as a unary string: the linear activation has no cost; square costs 2 bits; ReLU, sigmoid, and floor cost 4 bits; and the unit step function costs 8 bits.

\subsubsection{Connections and weights}

A connection's encoding includes its target unit number (each connection is specified within the description of its source unit, hence the source needs not be encoded), its weight, and its type: forward (0) or recurrent (1). 

To simplify the representation of weights in classical neural networks and to make it easier to mutate them in the genetic algorithm described below, we represent weights as signed fractions $\pm\frac{n}{d}$, which are serialized into bits by concatenating the codes for the sign (1 for $+$, 0 for $-$), the numerator and the denominator. For example, the weight $w_{ij} = +\frac{2}{5}$ would be encoded as:
\[
\underbrace{\underbrace{1}_{+}\underbrace{E(2)=10...10}_{2}\underbrace{E(5)=1110...11}_{5}}_{w_{ij}}
\]

\subsection{Search algorithm}

\begin{algorithm}[t]
\caption{Genetic algorithm}
\begin{algorithmic}

\State \textbf{function} \textsc{TournamentSelection}($pop$):
    \Indent
    \State $T \gets t$ random networks from $pop$
    \State $winner \gets argmin_{MDL}(T)$
    \State $loser \gets argmax_{MDL}(T)$
    \State \textbf{return} $winner$, $loser$
    \EndIndent
\State \textbf{end function}

\State $population \gets \emptyset$ \Comment{Population initialization}

\State \textbf{while} $|population| < N$ \textbf{do}:
\Indent
    \State generate a random network $net$
    \State add $net$ to $population$
\EndIndent
\State \textbf{end while}

\State $generation \gets 0$ \Comment{Evolution loop}
\State \textbf{while} $generation < Gen$ \textbf{do}:
\Indent
        \State \textbf{for} $N$ steps \textbf{do}:
    \Indent
        \State $parent, loser \gets $ \par $  \textsc{TournamentSelection}(population)$
        \State $\mathit{offspring} \gets mutate(parent)$
        \State $eval(\mathit{offspring})$ \Comment{MDL score}
        \State remove $loser$ from $population$
        \State add $\mathit{offspring}$ to $population$
    \EndIndent
\State \textbf{end for}
\State $generation \gets generation + 1$

\EndIndent
\State \textbf{end while}

\State \textbf{return} $argmin_{MDL}(population)$
\end{algorithmic}
\label{algorithm:basic_ga}
\end{algorithm}



Our interest in this paper is the MDL objective function and not the training method. However, identifying the MDL-optimal network is hard: the space of possible networks is much too big for an exhaustive search, even in very simple cases. We therefore need to combine the objective function with a suitable search procedure. We chose to use a genetic algorithm (GA; \citealp{Holland:1975}), which frees us from the constraints coming from backpropagation and is able to optimize the network structure itself rather than just the weights of a fixed architecture. For simplicity and to highlight the utility of the MDL metric as a standalone objective, we use a vanilla implementation of GA, summarized in Algorithm~\ref{algorithm:basic_ga}.

The algorithm is initialized by creating a population of $N$ random neural networks. A network is initialized by randomizing the following parameters: activation functions, the set of forward and recurrent connections, and the weights of each connection. Networks start with no hidden units. In order to avoid an initial population that contains mostly degenerate (specifically, disconnected) networks, output units are forced to have at least one incoming connection from an input unit.

The algorithm is run for $Gen$ generations, where each generation involves $N$ steps of selection followed by mutation. During selection, networks compete for survival on to the next generation based on their fitness, i.e., their MDL score, where lower MDL is better. 
A selected network is then mutated using one of the following operations: add/remove a unit; add/remove a forward or recurrent connection; add/remove a bias; mutate a weight or bias by changing its numerator or denominator, or flipping its sign; and change an activation function. These mutations make it possible to grow networks and prune them when necessary, and to potentially reach any architecture that can be expressed using our building blocks. The mutation implementations are based on~\citet{stanley_evolving_2002}.\footnote{A mutation step can potentially produce a network that contains loops in its non-recurrent connections, most commonly after a new connection is added. In the feed-forward phase, we detect loop-closing connections (using depth-first search) and ignore them. This avoids circular feeding, and at the same time creates a smoother search space, in which architectures freely evolve, even through intermediate defective networks. Stagnant loop connections which don't end up evolving into beneficial structures are eventually selected out due to the $|G|$ term.}

On top of the basic GA we use the Island Model (\citealp{gordon1993serial,Adamidis:1994,cantu1998survey}) which divides a larger population into `islands' of equal size $N$, each running its own GA as described above, periodically exchanging a fixed number of individuals through a `migration` step. This compartmentalization serves to mitigate against premature convergence which often occurs in large populations. The simulation ends when all islands complete $Gen$ generations, and the best network from all islands is taken as the solution. 

\section{Experiments}

We ran tasks based on several classical formal-language learning challenges. We use both deterministic and probabilistic expectations to test the ability of a learner to work on probabilistic and symbolic-like predictions. In addition to showing that the MDL learner performs well on test sets, we provide proofs that it performs well on the whole infinite language under consideration.

\subsection{General setup and comparison RNNs}
\label{sec:general_setup}

All simulations reported in this paper used the following hyper-parameters: 250 islands, each with population size 500 (total 125,000 networks), 25,000 generations, tournament size 2, migration size 2, and a migration interval of 30 minutes or 1,000 generations (earliest of the two). The number of generations was chosen empirically to allow enough time for convergence. Each task was run three times with different random seeds.\footnote{All experimental material and the source code for the model are available at \url{https://github.com/taucompling/mdlrnn}.}

To compare the performance of the MDL-optimized recurrent neural networks (MDLRNNs) with classical models, we trained standard RNNs on the same tasks, varying their architecture --- GRU \cite{cho_learning_2014}, LSTM \cite{hochreiter_long_1997}, and Elman networks \cite{Elman:1990} --- as well as the size of their hidden state vectors (2, 4, 32, 128), weight regularization method (L1/L2/none), and the regularization constant in case regularization was applied ($\lambda = 1.0/0.1/0.01$), totaling 84 RNN configurations. Each configuration was run three times with different random seeds.
These RNNs were trained with a cross-entropy loss, which corresponds to the $|D:G|$ term divided by the number of characters in the data.\footnote{All RNNs were trained using the Adam optimizer \citep{Kingma2015} with learning rate 0.001, $\beta_1 = 0.9$, and $\beta_2 = 0.999$. The networks were trained by feeding the full batch of training data for 1,000 epochs. The cross-entropy loss for RNNs is calculated using the natural logarithm, converted in Table~\ref{table:formal_lang_results} to base 2 for comparison with MDL scores.} 

Table~\ref{table:formal_lang_results} summarizes the results for both MDLRNNs and classical RNNs for all the tasks that will be described below. For each task, the representative network for each model (out of all configurations and random seeds) was chosen based on performance on the test set, using MDL scores for MDLRNNs and cross-entropy for RNNs.

It should be noted that model selection based on test performance is at odds with the premise of MDL: by balancing generalization and data fit during training, MDL automatically delivers a model which generalizes well beyond the training data; MDL also does away with the post-hoc, trial-end-error selection of hyper-parameters and regularization techniques inherent in classical models. In other words, MDL models can just as well be selected based on training performance. This is in contrast to standard RNNs, for which the training-best model is often the one that simply overfits the data the most. We show that even when given post-hoc advantage, RNNs still underperform.\footnote{When models are selected based on training performance (and then evaluated on the test sets), MDLRNNs outperform standard RNNs in all tasks in terms of cross-entropy and accuracy. We make the full training-based comparison available as part of the experimental material.}\textsuperscript{,}\footnote{Training-based selection yields different MDLRNN winners for three out of the seven relevant tasks when trained on the smaller data sets, and for two tasks when trained on the larger sets. However, only one of these cases, for $a^nb^nc^n$ with the larger training set, results in a drop from 100\% accuracy when selected by test to a suboptimum (97.6\%), while other models remain at the same accuracy levels.
MDL optimization is thus not immune to overfitting, which could occur for example due to accidental bad sampling. However, as made visible by our results, MDL training produces models that generalize well across data sets.}

\begin{table*}[]
\footnotesize

\begin{tabularx}{\textwidth}{>{\footnotesize}Yc|cc>{}c|cc|lc|c}
    \hline
    &
     Training&
    \multicolumn{3}{c|}{ Test cross-entropy ($\times 10^{-2}$) }&
    \multicolumn{2}{c|}{ Test accuracy (\%)} &
    \multicolumn{2}{c|}{ Best RNN} &
     MDLRNN\\ 
    &
    set size &
    MDLRNN & RNN & optimal &
    MDLRNN & RNN &
    Type & Size &
    proof\\
    \hline
\multirow{2}{*}{$a^nb^n$}
    & 100    & \textbf{29.4} & 53.2 & 25.8     & \textbf{100.0} & 99.8   & Elman & 2    &  \multirow{2}{*}{Th.~\ref{theorem:anbn}}   \\
    & 500    & \textbf{25.8} & 51.0 & 25.8     & \textbf{100.0} & 99.8    & Elman & 2    &    \\ \hline

\multirow{2}{*}{$a^nb^nc^n$} 
    & 100    & \textbf{49.3} & 62.6 & 17.2     & 96.5 & \textbf{99.8}    & Elman & 4    &  \multirow{2}{*}{Th.~\ref{theorem:an_bn_cn}}  \\
    & 500    & \textbf{17.2} & 55.4 & 17.2     & \textbf{100.0} & 99.8    & Elman & 4    &    \\ \hline

\multirow{2}{*}{$a^nb^nc^nd^n$} 
    & 100    & \textbf{65.3} & 68.1 & 12.9     & 68.6 & \textbf{99.8}    & GRU & 4    &     \\
    & 500    & \textbf{13.5} & 63.6 & 12.9     & \textbf{99.9} & 99.8    & GRU & 4    &     \\ \hline

\multirow{2}{*}{$a^nb^{2n}$}
    & 100    & \textbf{17.2} & 38.0 & 17.2     & \textbf{100.0} & 99.9    & Elman & 4    &  \multirow{2}{*}{Th.~\ref{theorem:an_b2n}}   \\
    & 500    & \textbf{17.2} & 34.7 & 17.2     & \textbf{100.0} & 99.9    & GRU & 4    &    \\ \hline
   
\multirow{2}{*}{$a^nb^mc^{n+m}$}
    & 100    & \textbf{39.8} & 47.6 & 26.9     & \textbf{98.9} & \textbf{98.9}    & Elman + L1 & 128    &  \multirow{2}{*}{Th.~\ref{theorem:an_bn_cnplusm}}   \\
    & 500    & \textbf{26.8} & 45.1 & 26.9     & \textbf{100.0} & 98.9    & Elman & 128    &    \\ \hline

\multirow{2}{*}{Dyck-1}
    & 100    & 110.7 & \textbf{94.5} & 88.2     & \textbf{69.9} & 10.9  & Elman & 4  & \multirow{2}{*}{Th.~\ref{theorem:dyck1}}    \\
    & 500    & \textbf{88.7} & 93.0 & 88.2     & \textbf{100.0} & 10.8     & LSTM & 4    &  \\ \hline

Dyck-2 \phantom{\raisebox{-6pt}{x}\raisebox{6pt}{x}}
    & 20,000    & \textbf{1.19} & \textbf{1.19} & 1.18     & \textbf{99.3} & 89.0     & GRU & 128    &     \\ \hline

\multirow{2}{*}{Addition}
    & 100    & \textbf{0.0} & 75.8 & 0.0     & \textbf{100.0} & 74.9     & Elman & 4    &   \multirow{2}{*}{Th.~\ref{proof:addition_net}}  \\
    & 400    & \textbf{0.0} & 72.1 & 0.0     & \textbf{100.0} & 79.4     & Elman & 4    &    \\ \hline

\end{tabularx}

\caption{Performance of the networks found by the MDL model compared with classical RNNs for the tasks in this paper. 
Test accuracy indicates \textit{deterministic accuracy}, the accuracy restricted to deterministic steps; Dyck-n tasks have no deterministic steps, hence here we report \textit{categorical accuracy}, defined as the fraction of steps where a network assigns a probability lower than $\epsilon = 0.005$ to each of the illegal symbols. 
When available, the last column refers to an infinite accuracy theorem for MDL networks: describing their behavior not only for a finite test set but over the relevant, infinite language.}

\label{table:formal_lang_results}
\end{table*}

\subsection{General measures}

We will report on two measures: (i)~\textbf{Accuracy}, which we explain below based on each task's unique properties; (ii)~\textbf{Cross-entropy}, averaged by character for better comparability across set sizes, and compared with a baseline value calculated from the probability distribution underlying each task. Most originally, in some occasions we report on these measures on whole, infinite languages. This is possible because, by design, MDL-optimized networks are just large enough for their task, allowing us to fully understand their inner workings.

\subsection{Experiment I: counters and functions}
\label{sec:experiment1_counters}

We test our model's ability to learn formal languages that require the use of one or multiple counters: $a^nb^n$, $a^nb^nc^n$, $a^nb^nc^nd^n$. These languages can be recognized using unbounded counting mechanisms keeping track of the number $n$ of $a$'s and balance the other characters accordingly.
We also test our model's ability to learn languages which not only encode integers as above, but also operate over them: $a^nb^mc^{n+m}$ (addition) and $a^nb^{2n}$ (multiplication by two). We did not test general multiplication through the language $a^nb^{m}c^{nm}$ for practical reasons, namely that the length of the sequences quickly explodes.

\subsubsection{Language modeling}
\label{sec:language_modeling}

The learner is fed with the elements from a sequence, one input after the next, and at each time step its task is to output a probability distribution for the next character. 
Following \citet{gers_lstm_2001}, each string starts with the symbol $\#$, and the same symbol serves as the target prediction for the last character in each string.

If the vocabulary contains $n$ letters, the inputs and outputs are one-hot encoded over $n$ input units (in yellow in the figures), and the outputs are given in $n$ units (in blue). To interpret these $n$ outputs as a probability distribution we zero negative values and normalize the rest to sum to $1$. In case of a degenerate network that outputs all 0's, the probabilities are set to the uniform value $1/n$. 

\subsubsection{Setup}

Each task was run with data set sizes of 100 and 500. The training sets were generated by sampling positive values for $n$ (and $m$, when relevant) from a geometric distribution with $p=.3$. The maximal values $K$ observed for $n$ and $m$ in our batches of size 100 and 500 were 14 and 22, respectively.

We test the resulting networks on all unseen sequences for $n$ in $[K+1,K+1001]$. For $a^nb^mc^{n+m}$ we test on $n$ and $m$ in $[K+1,K+51]$, i.e., the subsequent 2,500 unseen pairs. 

Only parts of the sequences that belong to the formal languages presented here can be predicted deterministically, e.g., for $a^nb^n$, the deterministic parts are the first $a$ (assuming $n>0$), all $b$'s except the first one, and the end-of-sequence symbol. For each of the tasks in this section, then, we report a metric of \textbf{deterministic accuracy}, calculated as the number of matches between the output symbol predicted with maximum probability and the ground truth, relative to the number of steps in the data that can be predicted deterministically.

\subsubsection{Results}

\begin{figure}[h]
\includegraphics[scale=0.18]{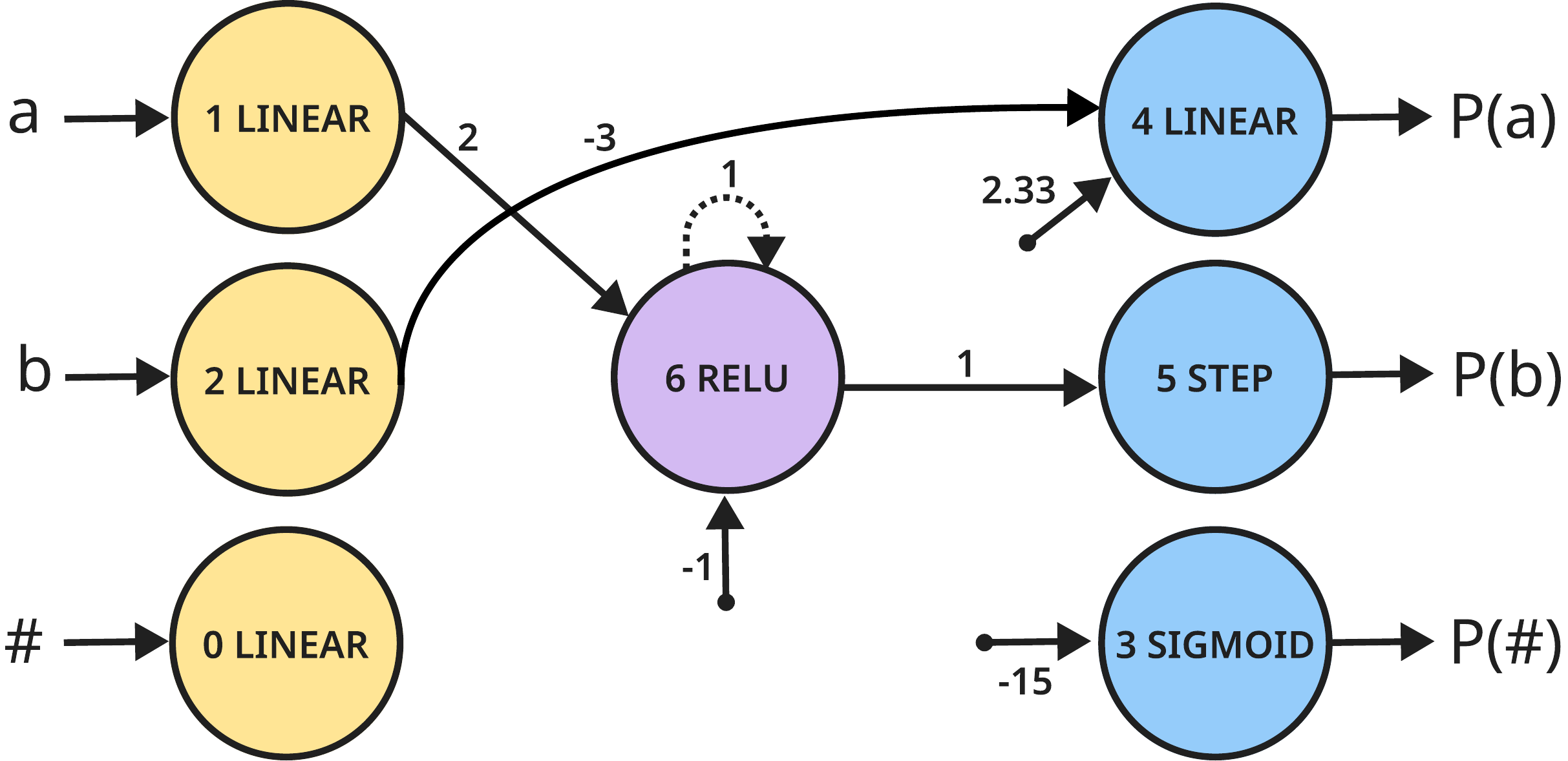}
\caption{The network found by the MDL learner for the $a^nb^n$ task, for a training set with data set size 500. See Theorem~\ref{theorem:anbn} for a description of how this network accepts any $a^nb^n$ sequence and why it rejects any other sequence.}
\label{fig:anbn_net}
\end{figure}

\begin{figure}[h]
\includegraphics[scale=0.185]{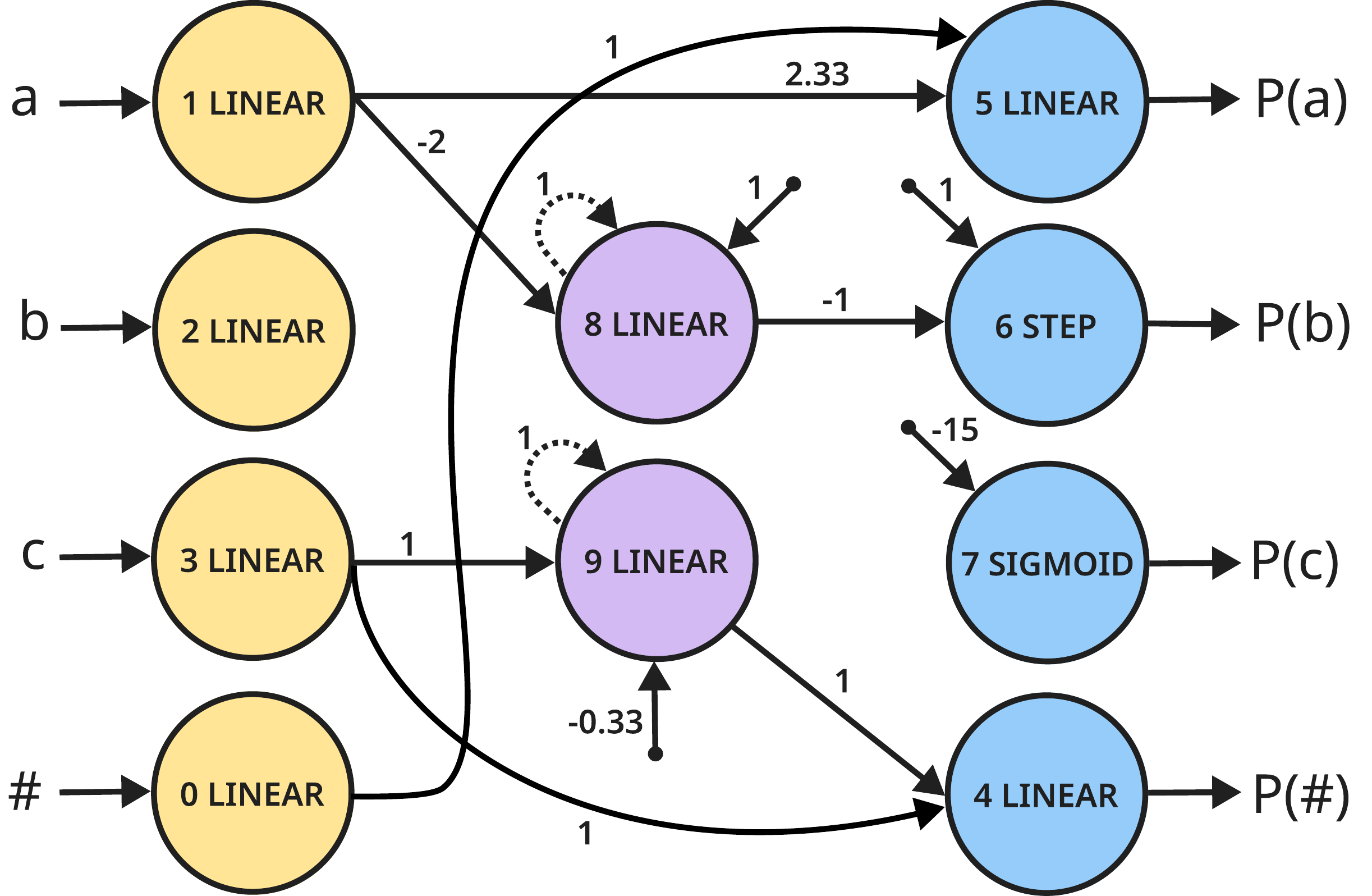}

\caption{The network found for the $a^nb^nc^n$ task for the larger training set. See Theorem~\ref{theorem:an_bn_cn} for a description of how this network accepts only sequences of the language $a^nb^nc^n$.}

\label{fig:anbncn_net}
\end{figure}

\begin{figure}[h]
\centering
\includegraphics[scale=0.185]{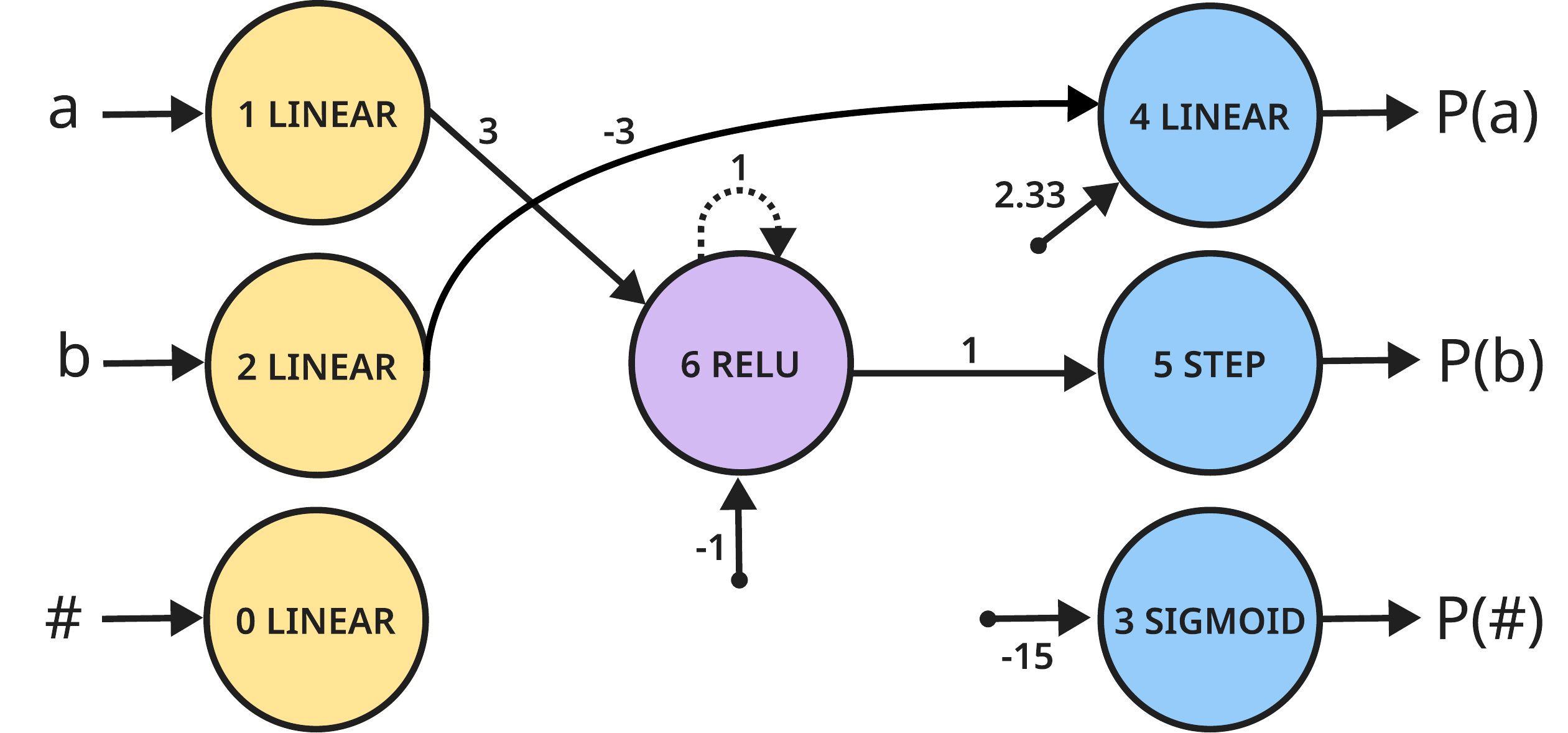}
\caption{The network found for the $a^nb^{2n}$ task for the larger training set. See Theorem~\ref{theorem:an_b2n} for a description of how this network accepts only sequences of the language $a^nb^{2n}$.}
\label{fig:anb2n_net}
\end{figure}

\begin{figure}[h]
\centering
\includegraphics[scale=0.185]{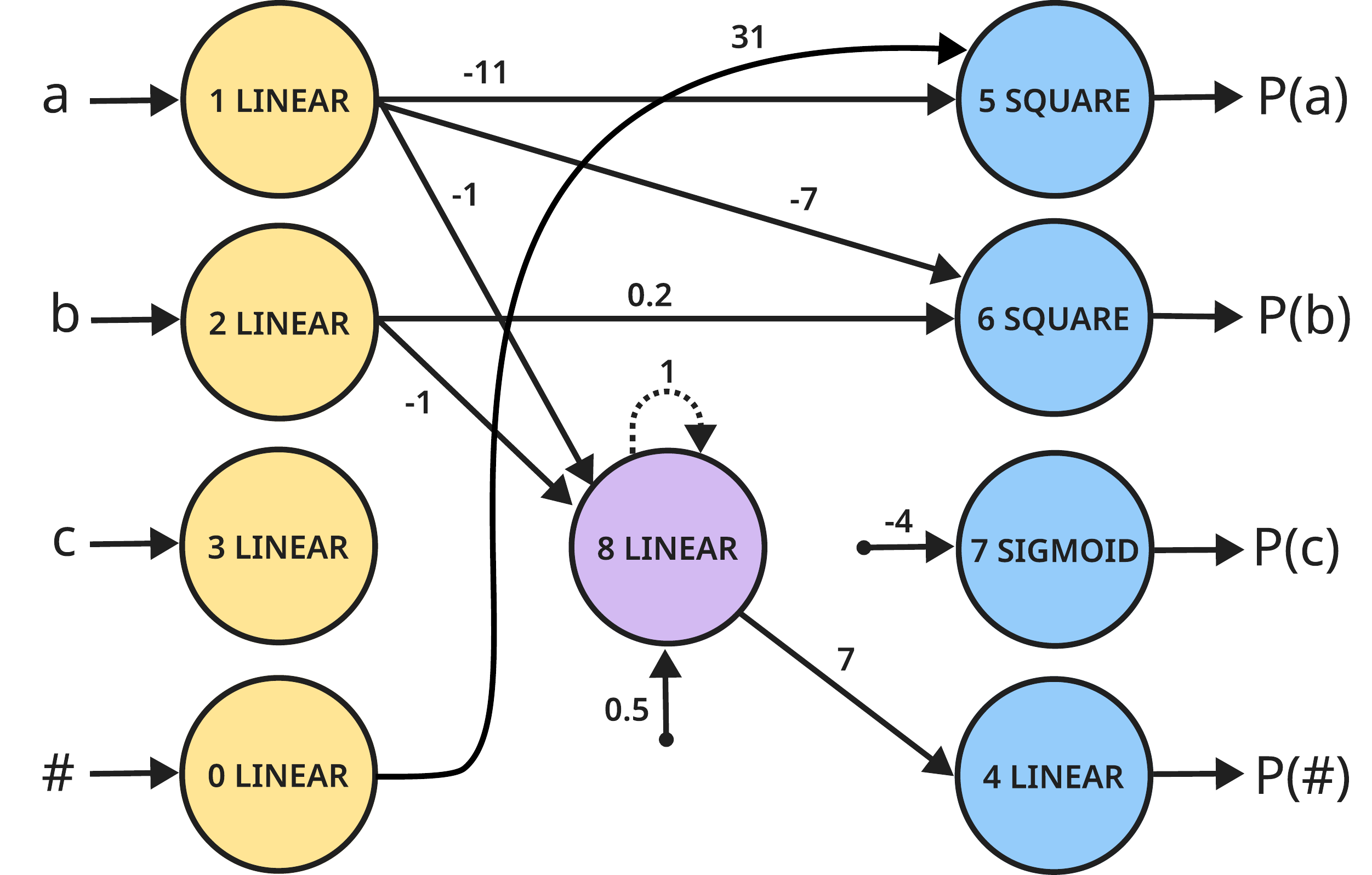}
\caption{The network found for the $a^nb^mc^{n+m}$ task for the larger training set. See Theorem~\ref{theorem:an_bn_cnplusm} for a  description  of  how  this  network  accepts only sequences of the language $a^nb^mc^{n+m}$.}

\label{fig:anbmcnplusm_net}
\end{figure}

The performance of the resulting networks is presented in Table~\ref{table:formal_lang_results}. In Figures~\ref{fig:anbn_net}-\ref{fig:anbmcnplusm_net}, we show the networks that were found and their typical behavior on language sequences. Thanks to their low number of units and connections, we are able to provide simple walkthroughs of how each network operates. We report the following measures:

\paragraph{Deterministic accuracy: perfect for almost all tasks, both with small and large training sets.}
The MDL learner achieves perfect accuracy for the tasks $a^nb^n$ and $a^nb^{2n}$, both with small and large training sets. The learner also achieves perfect accuracy for $a^nb^nc^n$ and $a^nb^mc^{n+m}$ with a larger training set, and in fact the networks found there would be better optima also for the respective smaller training sets, therefore showing that the suboptimal results for the small training sets are only due to a limitation of the search, and that perfect accuracy should in principle be reachable there too with a more robust search. 

The only task for which MDLRNNs did not reach 100\% accuracy is $a^nb^nc^nd^n$. Since the other tasks show that our representations make it possible to evolve counters, we attribute this failure to the search component, assuming a larger population or more generations are needed, rather than lack of expressive power; networks for this task require more units for inputs and outputs, which enlarge the number of possible mutations the search can apply at each step.

\paragraph{Cross-entropy: near perfect.}
For all tasks but $a^nb^nc^nd^n$, the MDLRNN per-character average cross-entropy is also almost perfect with respect to the optimal cross-entropy calculated from the underlying probability distribution. 

\paragraph{RNNs: no perfect generalization.}
Among the competing models, no standard RNN reached 100\% deterministic accuracy on the test sets, and all RNNs reached suboptimal cross-entropy scores, indicating that they failed to induce the grammars and probability distributions underlying the tasks. 
In terms of architecture size, the best-performing RNNs are often those with fewer units, while L1 and L2 regularizations do not yield winning models except for one task.

\paragraph{Transparency supports formal proofs that results are perfect for the whole, infinite language.}
For all tasks but $a^nb^nc^nd^n$ then, deterministic accuracy and cross-entropy are perfect/excellent on training and test sets. Because the MDL networks are small and transparent, we can go beyond these results and demonstrate formally that the task is performed perfectly on the entire infinite underlying language. To our knowledge, such results have never been provided for any classic neural network in these tasks or any other.

\begin{theorem}
\label{theorem:anbn}
The $a^nb^n$ network represented in Fig.~\ref{fig:anbn_net} outputs the correct probability for each character, for each sequence in the $a^nb^n$ language, with a margin of error below $10^{-6}$.
\end{theorem}

\begin{table}[h]
\scriptsize

\begin{tabularx}{\columnwidth}{Y|rrrr}
$a^nb^n$ &
    Unit 6 & Unit 4 & Unit 5 & Unit 3 \\
    &
    & $\boldsymbol{P(a)}$ & $\boldsymbol{P(b)}$ & $\boldsymbol{P(\#)}$ \\
\hline
Initial \# &
    0 & 7/3 & 0 & \tiny$\sigma(-15)$ \\
         &
     & $\boldsymbol{\sim\!1}$ & $\boldsymbol{\sim\!0}$ & $\boldsymbol{\sim\!0}$ \\
\hline
$k^{\textrm{th}}$ a &
    $k$ & 7/3 & 1 & \tiny$\sigma(-15)$ \\
         &
     & $\boldsymbol{\sim\!.7}$ & $\boldsymbol{\sim\!.3}$ & $\boldsymbol{\sim\!0}$ \\
\hline
$k^{\textrm{th}}$ b, &
    $n\!-\!k$ & -2/3 & 1 & \tiny$\sigma(-15)$ \\
$k<n$         &
     & \textbf{0} & $\boldsymbol{\sim\!1}$ & $\boldsymbol{\sim\!0}$ \\
\hline
$n^{\textrm{th}}$ b &
    0 & -2/3 & 0 & \tiny$\sigma(-15)$ \\
         &
     & $\boldsymbol{0}$ & $\boldsymbol{0}$ & $\boldsymbol{1}$ 
\end{tabularx}
\caption{Unit values (columns) during each phase of a valid $a^nb^n$ sequence (rows). The second line for output units, given in bold, indicates the final normalized probability.}
\label{table:proof_table_anbn}
\end{table}

\begin{proof}
Table~\ref{table:proof_table_anbn} traces the value of each unit at each step in a legal sequence for the relevant network. When normalizing the outputs to obtain probabilities, the values obtained are the exact ground-truth values, up to the contribution of $\sigma(-15)$ to that normalization (sigmoid is abbreviated as $\sigma$),
which is negligible compared to all other positive values (the largest deviance is $\frac{\sigma(-15)}{1+\sigma(-15)}\approx 3.10^{-7}$, observed during the $b$'s). The network not only accepts valid $a^nb^n$ sequences, but also rejects other sequences, visible by the zero probability it assigns to irrelevant outputs at each phase in Table~\ref{table:proof_table_anbn}.

More informally, 
the network uses a single hidden unit (6) as a counter, recognizable from the recurrent loop onto itself. The counter is incremented by 1 for each $a$ (+2 from unit 1, -1 from the bias), and then decremented by 1 for each $b$ (signaled by a lack of $a$, which leaves only the -1 bias as input to the counter).
\end{proof}

\begin{theorem}
\label{theorem:an_bn_cn}
The network represented in Fig.~\ref{fig:anbncn_net} outputs the correct probability for each character, for each sequence in the $a^nb^nc^n$ language, with a margin of error below $10^{-6}$.
\end{theorem}

\begin{table}[h]
\scriptsize

\begin{tabularx}{\columnwidth}{l|@{ }r@{ }r@{ }r@{ }r@{ }r@{ }r}
$a^nb^nc^n$ &
    Unit 8 & Unit 9 & 
    Unit 4 & Unit 5 & Unit 6 & Unit 7 \\
    &
    &&
    $\boldsymbol{P(\#)}$ & $\boldsymbol{P(a)}$ & $\boldsymbol{P(b)}$ & $\boldsymbol{P(c)}$ \\
\hline
Initial $\#$ &
    1 & 
    $-\frac{1}{3}$ & 
    $-\frac{1}{3}$ & 
    1 & 
    0 & 
    \tiny$\sigma(\!-\!15\!)$ \\
    &
    & &
    $\boldsymbol{0}$ & $\boldsymbol{\sim\!1}$ & $\boldsymbol{0}$ & $\boldsymbol{\sim\!0}$ \\
\hline

$k^{\textrm{th}}$ a &
    $1-k$ &
    $-\frac{k+1}{3}$ & 
    $-\frac{k+1}{3}$ & $\frac{7}{3}$ & 
    1 & \tiny$\sigma(\!-\!15\!)$ \\
&
    & &
    $\boldsymbol{\sim\!0}$ & $\boldsymbol{\sim\!.7}$ & $\boldsymbol{\sim\!.3}$ & $\boldsymbol{\sim\!0}$ \\
\hline

$k^{\textrm{th}}$ b &
    $k\!+\!1\!-\!n$ & $-\frac{k+n+1}{3}$ & 
    $-\frac{k+n+1}{3}$ & 
    0 & 1 & \tiny$\sigma(\!-\!15\!)$ \\
$k<n$ &
    & &
    $\boldsymbol{0}$ & $\boldsymbol{0}$ & $\boldsymbol{\sim\!1}$ & $\boldsymbol{\sim\!0}$ \\
\hline

$n^{\textrm{th}}$ b &
    1 & 
    $-\frac{2n+1}{3}$ & 
    $-\frac{2n+1}{3}$ & 0 & 0 & \tiny$\sigma(\!-\!15\!)$ \\
 &
    & &
    $\boldsymbol{0}$ & $\boldsymbol{0}$ &$\boldsymbol{0}$ & $\boldsymbol{1}$ \\
\hline

$k^{\textrm{th}}$ c &
    $1\!+\!k$ & 
    $\frac{2k\!-2n-\!1}{3}$ & 
    $\frac{2(k\!+\!1\!-\!n)}{3}$ & 0 & 0 & \tiny$\sigma(\!-\!15\!)$ \\
$k<n$ &
    & &
    $\boldsymbol{0}$  & $\boldsymbol{0}$  & $\boldsymbol{0}$  & $\boldsymbol{1}$ \\
\hline

$n^{\textrm{th}}$ c &
    $1\!+\!n$ & $-\frac{1}{3}$ & 
    $\frac{2}{3}$ & 0 & 0 & \tiny$\sigma(\!-\!15\!)$ \\
 &
    & &
    $\boldsymbol{\sim\!1}$  & $\boldsymbol{0}$ & $\boldsymbol{0}$ & $\boldsymbol{\sim\!0}$  \\

\end{tabularx}
\caption{Unit values (columns) during each phase of a valid $a^nb^nc^n$ sequence (rows).}
\label{table:proof_table_anbcn}
\end{table}

\begin{proof}
\label{proof:an_bn_cn}
The proof is again obtained by tracing the values each unit holds at each phase of a valid sequence in the language, see Table~\ref{table:proof_table_anbcn}.

The network uses two hidden units that serve as counters for the number of $a$'s (unit 8) and $c$'s (unit 9). Each occurrence of $a$ simultaneously feeds the output unit for $a$ (5) and the $a$ counter (8) connected to the $b$ output (6), using weights to create the correct probability distribution between $a$'s and $b$'s. Once $a$'s stop, $P(a)$ flatlines, and the $a$ counter (8) starts decreasing until $n$ $b$'s are seen. Another counting system has evolved in unit 9 which counts the number of $a$'s and $b$'s (signaled by lack of $c$'s), and then decreases for each $c$, finally triggering the end-of-sequence output $\#$.
Note how the model avoids holding a third counter for the number of $b$'s, by reusing the $a$ counter. This makes it possible to disconnect the $b$ input unit (2), which minimizes encoding length.
\end{proof}

\begin{theorem}
\label{theorem:an_b2n}
The $a^nb^{2n}$ network represented in Fig.~\ref{fig:anb2n_net} outputs the correct probability for each character, for each sequence in the $a^nb^{2n}$ language, with a margin of error below $10^{-6}$.
\end{theorem}

\begin{proof}
The network is similar to the one found for $a^nb^n$ (Fig.~\ref{fig:anbn_net}). The proof that this network is accurate is also similar (Theorem~\ref{theorem:anbn}), the only difference being that the hidden unit is incremented with 2 instead of 1 for each $a$ input.
\end{proof}

\begin{theorem}
\label{theorem:an_bn_cnplusm}
The network represented in Fig.~\ref{fig:anbmcnplusm_net} outputs the correct probability for each character, for each sequence in the $a^nb^mc^{n+m}$ language, with a margin of error below $.2$ (and below $10^{-4}$ for deterministic steps, i.e., probabilities 0 or 1).\footnote{For this task, the average test cross-entropy per character of the network trained on the larger data set goes slightly below the optimum (see Table~\ref{table:formal_lang_results}); this can happen for example if the model picks up on unintentional regularities in the training set that are also present in the test set.}
\end{theorem}

\begin{table}[h]
\scriptsize
\begin{tabularx}{\columnwidth}{l|rrrr}
$a^nb^mc^{n+m}$ &
    unit 5 & unit 6 & unit 7 & unit 4 \\
    &
    $\boldsymbol{P(a)}$ & $\boldsymbol{P(b)}$ & $\boldsymbol{P(c)}$ & $\boldsymbol{P(\#)}$ \\
\hline
Initial $\#$ &
    $31^2$ & 0 & $\sigma(-4)$ & $\frac{7}{2}$\\
         &
    $\boldsymbol{\sim\!.996}$ & $\boldsymbol{0}$ & $\boldsymbol{\sim\!0}$ & $\boldsymbol{\sim\!.004}$\\
\hline
$k^{\textrm{th}}$ a &
    $11^2$ & $7^2$ & $\sigma(-4)$ & $\frac{7}{2}(1-k)$ \\
    &
    $\boldsymbol{\sim\!.71}$ & $\boldsymbol{\sim\!.29}$ & $\boldsymbol{\sim\!0}$ & $\boldsymbol{0}$ \\
\hline

$k^{\textrm{th}}$ b &
    0 & $.04$ & $\sigma(-4)$ & $\frac{7}{2}(1\!-\!n\!-\!k)$ \\
    &
    $\boldsymbol{0}$ & $\boldsymbol{\sim\!.69}$ & $\boldsymbol{\sim\!.31}$ & $\boldsymbol{0}$ \\
\hline

$k^{\textrm{th}}$ c &
    0 & 0 & $\sigma(-4)$ & $\frac{7}{2}(1\!-\!n\!-\!m\!+\!k)$ \\
$k<m+n$    &
    $\boldsymbol{0}$ & $\boldsymbol{0}$ & $\boldsymbol{1}$ & $\boldsymbol{0}$ \\
\hline

$(m+n)^{\textrm{th}}$ c &
    0 & 0 & $\sigma(-4)$ & $\frac{7}{2}$ \\
    &
    $\boldsymbol{0}$ & $\boldsymbol{0}$ & $\boldsymbol{\sim\!0}$ & $\boldsymbol{\sim\!1}$ 
    
\end{tabularx}
\caption{Unit values during each phase of a valid $a^nb^mc^{n+m}$ sequence.}
\label{table:proof_an_bm_cnplusm}
\end{table}

\begin{proof}
In Table~\ref{table:proof_an_bm_cnplusm} we trace the values of each unit during feeding of a valid sequence in $a^nb^mc^{n+m}$. We do not represent the internal memory unit 8, its value is the seventh of that of unit 4.

Here, a network with a single counter (unit 8) has evolved which recognizes the language with 100\% accuracy. While one would think that this task requires at least two counters --- for $n$ and $m$ --- the pressure for parsimony leads the model to evolve a more compact solution: since the number of $c$'s is always $n+m$, and no other symbols appear between the first and last $\#$, the network uses the signal of \textit{lack of} $c$'s as an indication of \textit{positive occurrences} of either $a$ or $b$. This might raise a suspicion that the network recognizes out-of-language sequences such as balanced yet unordered strings, e.g. $abbaaccccc$.
In practice, however, the network imposes a strict order: $a$ receives a positive probability only after $\#$ or $a$; $b$ only after $a$ or $b$; and $c$ receives a significant proportion of the probability mass only as a last resort.
\end{proof}

\subsection{Experiment II: Dyck-1 vs. Dyck-2}
\label{sec:experiment_dyck1-dyck2}

In previous tasks, we showed the capability of MDLRNNs to evolve counters. A counter is also what is needed to recognize the Dyck-1 language of well-matched parentheses sequences. In the Dyck-2 language, however, there are two pairs of opening and closing characters, such as parentheses and brackets. Counters are not sufficient then, and a stack is needed to additionally track whether the next closing element must be a parenthesis or a bracket (and similarly for any Dyck-n language for $n>1$, \citealp{suzgun_memory-augmented_2019}). We ask here whether MDL-optimized networks can evolve not only counters but also stacks.

\subsubsection{Setup}

The setup is that of a language modeling task, as in Experiment I. For Dyck-1, the training sequences were generated from a PCFG with probability $p=.3$ of opening a new bracket, with data set sizes 100 and 500. The test sets contained 50,000 sequences generated from the same grammar that were not seen during training.

For Dyck-2, a fully operational stack is needed in order to recognize the language. We thus first make sure that such a network exists in the search space. We do this by manually designing a network that implements a fully operational stack. 
We use this network as a baseline for comparison with the results of the MDL simulations. 

\begin{figure*}
\centering
\includegraphics[width=\textwidth]{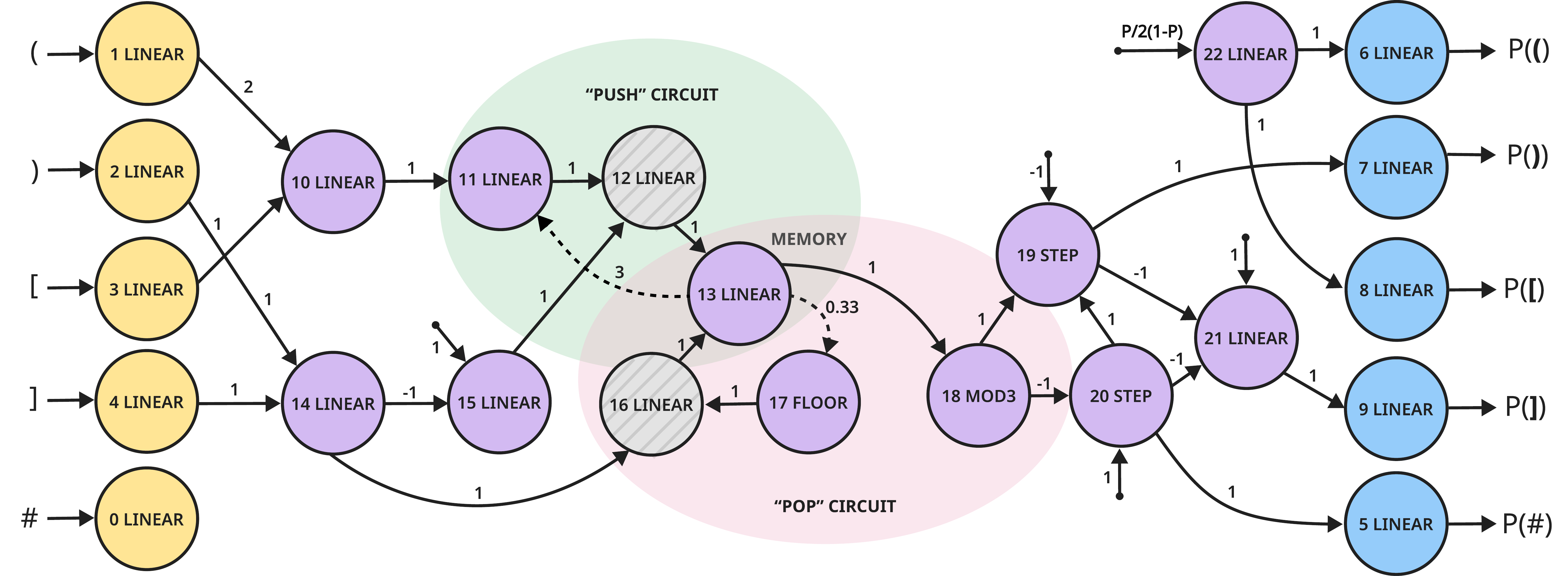}
\caption[width=\textwidth]{A manually-designed network implementing a fully operational stack, which recognizes the Dyck-2 language. The network uses an additional type of unit, which calculates the product of its inputs instead of summing them, making it possible to create gate units similar to those of LSTM networks (gray striped units in the figure). The stack's memory is implemented as an integer, stored here in unit 13; the integer is shifted to the left or right in base 3, making it possible to store the value 2 for a parenthesis and 1 for a bracket, visible in their respective input weights. Unit 12 is the `push' gate, which opens when a non-zero value flows from the opening bracket or parenthesis inputs. Unit 16 is the `pop' gate, opened by a non-zero input from a closing symbol. The recurrent connection from memory unit 13 to unit 11 performs the base-3 left shift by multiplying the memory by 3. For `pop', a right shift is applied by dividing the memory by 3. To extract the value of the topmost element, modulo 3 is applied. The bias for unit 22 handles outputting the probability $p$ of opening a new bracket/parenthesis.}
\label{fig:net_dyck2_manual}
\end{figure*}

The stack network and a detailed description of its mechanism are given in Fig.~\ref{fig:net_dyck2_manual}. 
We add two additional building blocks in order to implement this mechanism: the modulo 3 activation function used in the `pop' implementation, and a second type of unit which applies multiplication to its inputs, in order to create gates such as the ones used in LSTM networks.
Because the inclusion of these new representations enlarges the search space, and because the baseline network is larger in terms of number of units than the networks found in previous tasks (23 vs. 7-10), we double the genetic algorithm's overall population size (500 islands vs. 250), allowing more hypotheses to be explored. We also enlarge the training set to 20,000 samples, which allows networks with costlier $|G|$ terms to evolve. Here again the training sequences were generated from a PCFG with probability $p=.3$ for opening a new bracket or parenthesis, and tested on 50,000 novel sequences generated from the same grammar. 

Dyck sequences don't have any sub-parts which can be predicted deterministically (one can always open a new bracket), which makes \textit{deterministic accuracy} reported above irrelevant. We report instead a metric we call \textbf{categorical accuracy}, defined as the fraction of steps where the network predicts probability $p\ge\epsilon$ for symbols that \textit{could} appear at the next step, and $p<\epsilon$ for irrelevant symbols. For example, for Dyck-2, when the upcoming closing character is a bracket (i.e., the last seen opening character was a bracket), the network should assign probability 0 to the closing parenthesis; and similarly for the end-of-sequence symbol as long as a sequence is unbalanced.
Because classical RNNs cannot assign categorical 0 probabilities to outputs due to their reliance on softmax layers, we use $\epsilon = 0.005$ as a categorical margin.

\subsubsection{Results}
\label{sec:dyck_results}

Full performance details are given in Table~\ref{table:formal_lang_results}.

For the Dyck-1 language, the networks for the small and large training sets reach average test cross-entropy of 1.11 and 0.89 respectively, compared to an optimal 0.88. This result is in line with those of Experiment I, where we have shown that our representations are capable of evolving counters, which are sufficient for recognizing Dyck-1 as well. An Elman RNN reaches a better cross-entropy score, but worse categorical accuracy, for the smaller training set, while MDLRNN wins with the larger set, reaching a score close to the optimum and 100\% categorical accuracy. 

\begin{figure}[h]
\centering
\includegraphics[scale=0.187]{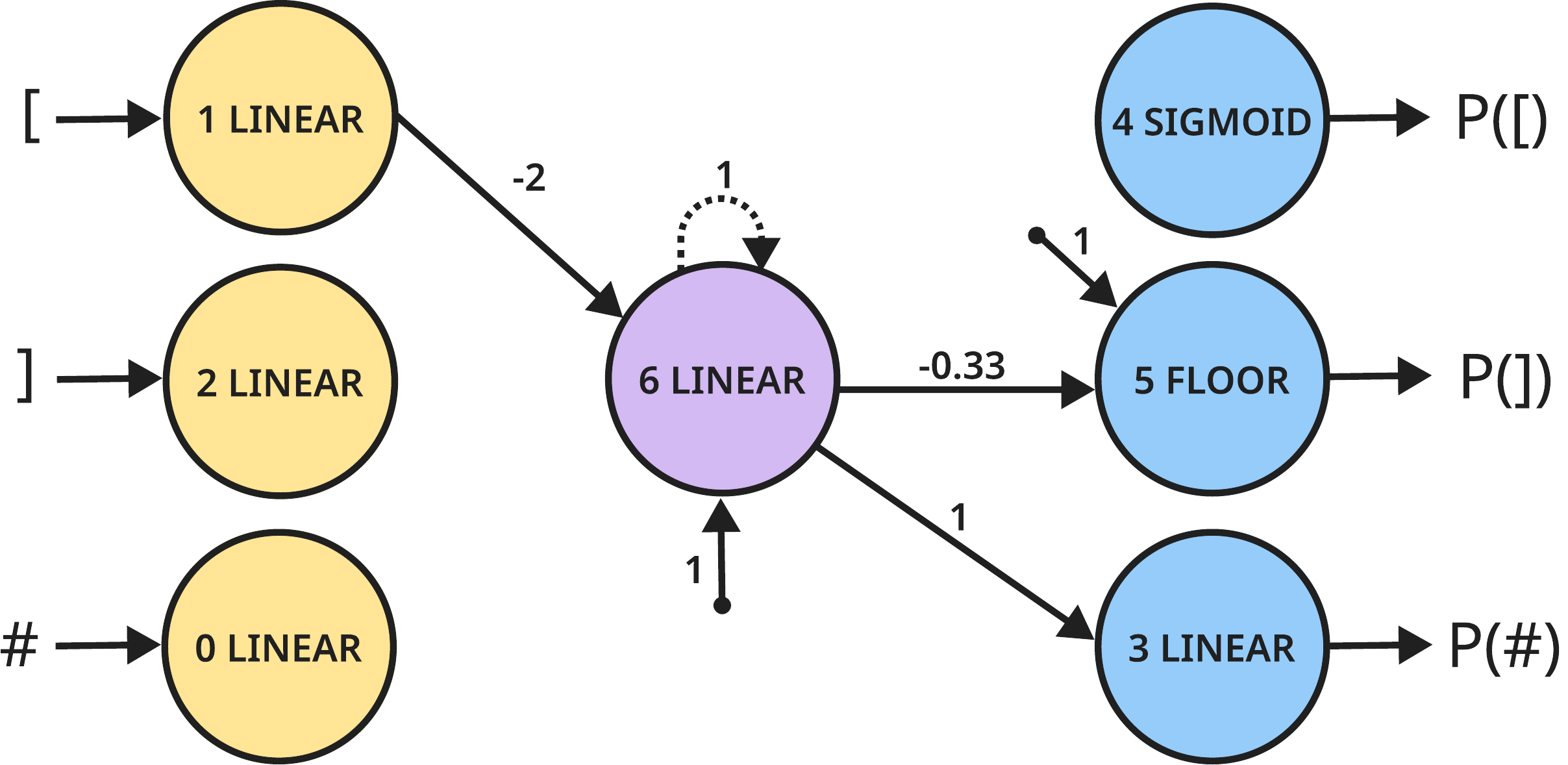}
\caption{The network found by the MDL learner for the Dyck-1 task for the larger training set. See Theorem~\ref{theorem:dyck1} for a description of how it accepts only valid Dyck-1 sequences.
}
\label{fig:net_dyck1}
\end{figure}

\begin{theorem}
\label{theorem:dyck1}
When brackets are well balanced, the Dyck-1 network in Fig.~\ref{fig:net_dyck1} correctly predicts that no closing bracket can follow by assigning it probability 0. Conversely, when brackets are unbalanced, it assigns probability 0 to the end-of-sequence symbol.
\end{theorem}

\begin{table}[h]
\scriptsize

\begin{tabularx}{\columnwidth}{Y|rrrr}
Dyck-1 &
    Unit 6 & Unit 4 & Unit 5 & Unit 3 \\
    &
    & $\boldsymbol{P([)}$ & $\boldsymbol{P(])}$ & $\boldsymbol{P(\#)}$ \\
\hline

$o>0$ &
    $1-o$ & $1\!/2$ & $floor(\frac{2+o}{3})$ & $1-o$ \\
         &
     & $\boldsymbol{\sim\!\frac{3}{7+2o}}$ & $\boldsymbol{\sim\!\frac{4+2o}{7+2o}}$ & $\boldsymbol{0}$ \\

\hline

$o=0$ &
    $1$ & $1\!/2$ & $floor(\frac{2}{3})=0$ & $1$ \\
         &
     & $\boldsymbol{1\!/3}$ & $\boldsymbol{0}$ & $\boldsymbol{2\!/3}$ \\

\end{tabularx}
\caption{Unit values and output probabilities during the two possible phases of a Dyck-1 sequence: (i) the number of open brackets $o$ is positive, or (ii) all brackets are well balanced ($o=0$).}
\label{table:proof_table_dyck1}
\end{table}

\begin{proof}
Call $o$ the number of open brackets in a prefix.
Throughout a Dyck-1 sequence, unit 6 holds the value $1-o$: it holds the value $1$ after the initial `\#'; then $+1$ is added for each `$[$', and $-1$ for each `$]$'.
The output probabilities in the cases of balanced and unbalanced sequences are then given in Table~\ref{table:proof_table_dyck1}. 
The theorem follows from the fact that $P(\#)=0$ if $o>0$, and $P(])=0$ if $o=0$.
In the target language, we note that opening brackets have a constant probability of $P([)=.3$, while in the found network this probability decreases with $o$ (visible in unit 4's output probability, Table~\ref{table:proof_table_dyck1}).
This makes a potential difference for high values of $o$, which however are very rare 
($o$ decreases with probability $.7$ at all time steps).
\end{proof}

For Dyck-2, the MDL model fails to reach the architecture of the baseline manual network, or another architecture with a similar cross-entropy score, reaching a network which has a worse MDL score than the baseline (148,497 vs. 147,804).
Accordingly, MDLRNN reaches a non-perfect 99.27\% categorical accuracy, compared to 89.01\% for RNNs, which reflects both models' failure to correctly balance certain sequences. Both models tie at 1.19 cross-entropy, close to the optimal 1.18.

Since we confirmed that the baseline architecture exists in the search space, we conclude that reaching a fully operational stack network is hindered by the non-exhaustive search procedure, rather than by the MDL metric. 
This may be solvable by tweaking the hyper-parameters or putting more computational resources into the search.
It could be, however, that this difficulty is due to a more interesting property of the task at hand. It has been claimed that evolutionary algorithms tend to struggle with so-called `deceptive' optimization problems --- tasks for which series of intermediate good solutions don't necessarily lead to a global optimum (see overview in \citealp{lehman_abandoning_2011}). For the stack network, it could be the case that a stack is only operational in its full form, and that intermediate networks deceive and lead the search to local minima, like the one found in the current simulation. 

A recent line of work has addressed the need for stacks by manually designing stack-like mechanisms using continuous representations, and integrating them manually into standard architectures (\citealp{graves_neural_2014}, \citealp{joulin_inferring_2015}, \citealp{suzgun_memory-augmented_2019}, among others). Indeed, when they are explicitly augmented with manually-designed continuous stack emulators, neural networks seem to be able to capture nonregular grammars such as the one underlying the Dyck-2 language. Similarly, we could allow our search to add stacks in one evolution step. This could overcome the risk of a deceptive search target mentioned above. If successful, we can expect this approach to come with all the benefits of the MDL approach: the winning network would remain small and transparent, and it would eventually contain a memory stack only if this is intrinsically needed for the task.

\subsection{Experiment III: general addition}
\label{sec:experiment3_addition}

In the previous experiments, we saw that MDL-optimized networks are capable of representing integers and add them in what amounts to unary representation (see $a^nb^mc^{n+m}$ language). Here, we show that addition can be performed when the numbers and outputs are represented in a different format. Specifically, we consider the familiar task of adding two integers in binary representation when the numbers are fed bit-by-bit in parallel, starting from the least significant bit. While this problem has good approximate solutions in terms of standard RNNs,%
\footnote{An example implementation that reportedly works up to a certain number of bits: \url{https://github.com/mineshmathew/pyTorch_RNN_Examples}
}
we will show that our model provides an \emph{exact} solution. As far as we are aware, this has not been shown before.

\subsubsection{Setup}

In this setting, we diverge from a language modeling task. The network here is fed at each time step $i$ with a tuple of binary digits, representing the digits $n_i$ and $m_i$ of two binary numbers $n$ and $m$, starting from the least significant bit. The two input units are assigned the values $n_i$ and $m_i$. The output is interpreted as the predicted probability that $(n+m)_i=1$, that is that $1$ is the $i^\textrm{th}$ digit in the sum $(n+m)$. Output values are capped to make them probabilities: values at or below $0$ are interpreted as probability 0, values at or above 1 are interpreted as probability~1.

The model was trained on two corpus sizes: one that contained all pairs of integers up to $K=10$ (total 100 samples), and a larger set of all pairs up to $K=20$ (total 400). The resulting networks were then tested on the set of all pairs of integers $n, m \in [K+1,K+251]$, i.e., 62,500 pairs not seen during training. Since the task is fully deterministic, we report a standard accuracy score.

\subsubsection{Results}

\begin{figure}[h]
\centering
\includegraphics[scale=0.185]{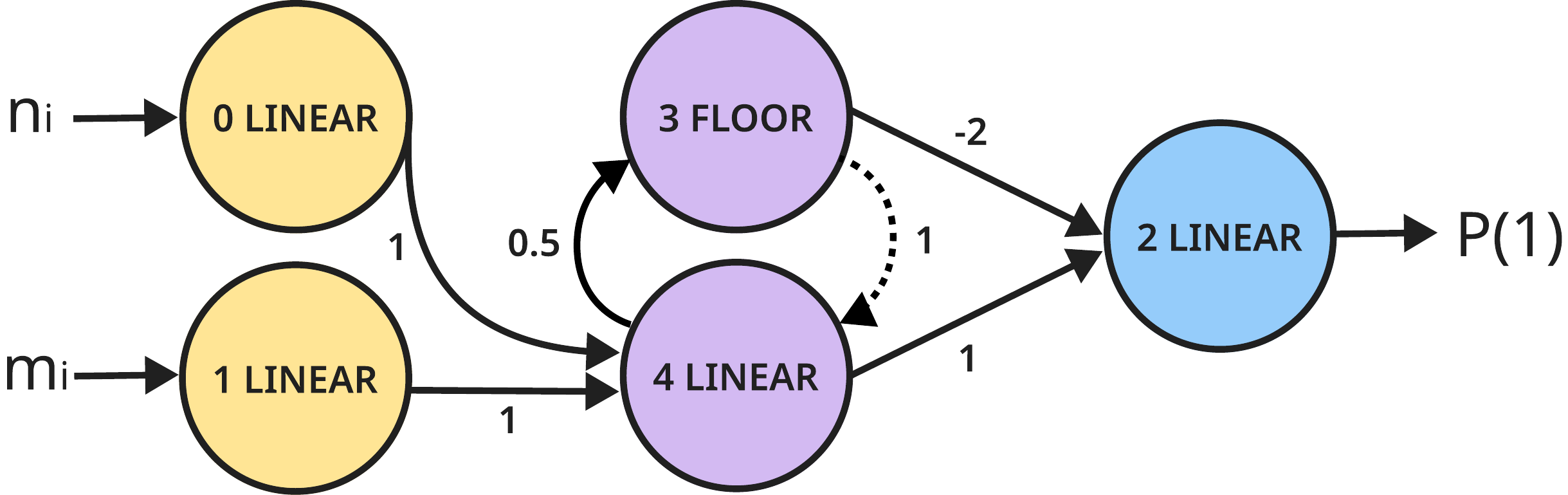}
\caption{The network found by the MDL learner for the binary addition task, trained on all 400 pairs of numbers up to 20. 
This network is correct for all numbers (Theorem~\ref{proof:addition_net}).}.

\label{fig:MDLnet-addition}
\end{figure}

MDLRNNs reached 100\% accuracy on both test sets, and an optimal cross-entropy score of zero.  
Fig.~\ref{fig:MDLnet-addition} shows the MDLRNN result for the larger training set. It provably does perfect addition, with perfect confidence, for all pairs of integers:

\begin{theorem}
\label{proof:addition_net}
For the net in Fig.~\ref{fig:MDLnet-addition}, the output unit at time step $i$ is the $i^\textrm{th}$ digit of the sum of the inputs.
\end{theorem}

\begin{proof}
Call $c_{i-1}^3$ the value of unit 3 at time step $i-1$; this value is the carry-over for the next time step, feeding unit 4 through their recurrent connection at time step $i$.
This can be proven in two steps. (1)~At the first time step $i=1$ the carry-over going into unit 4 is $0$, since recurrent inputs are $0$ by default at the first time step. (2)~By induction, $c_i^4$ is the sum of the relevant carry-over ($c_{i-1}^3$) and the two input digits at time~$i$. The combination of the $1/2$ multiplication and floor operation extracts a correct carry-over value from that sum and stores it in unit 3.
From there, we see that $c_i^2$ holds the correct binary digit: the sum of current inputs and carry-over (from $c_i^4$), minus the part to be carried over next (from $-2 \times c_i^3$).
\end{proof}

Again, the task is learned perfectly and in a readable fashion. As a side remark, the network obtained here can also naturally be extended to perform addition of more than 2 numbers, simply by adding the necessary inputs for the additional digits and connecting them to cell 4. To our knowledge no other RNN has been proven to hold a carry-over in memory for an unbounded number of digits, i.e.~to perform general addition of any arbitrary pair of numbers. The best competing classical RNNs trained here were never able to reach more than 79.4\% accuracy on the test sets, indicating that they learned a non-general way to do addition.

\subsection{Objective function probe}

\label{sec:partial_mdl}
In order to further probe the value of the MDL objective function --- and to isolate the effects of the objective function, which is our main focus, from those of the training method and the activation functions --- we ran four additional simulations using variations of MDL while keeping the setting without change. The variants of the objective function that we tested are: (i) $|G|$ alone, i.e., only the description length of the network is minimized; (ii)
~$|D:G|$ alone, i.e., the model only optimizes training data fit, similarly to a cross-entropy loss in traditional models; (iii)-(iv)~replacing $|G|$ with traditional L1 and L2 regularization terms.

The different objective functions were tested on the $a^nb^n$ task using the same hyper-parameters given in Sec.~\ref{sec:general_setup}. Table~\ref{table:partial_mdl} summarizes the performance for each resulting network. As expected, when $|G|$ alone is minimized, the result is a degenerate network with no hidden units or connections.
Conversely, $|D:G|$-only training results in a network growing large and picking up on accidental regularities in the training set. The overfitting leads to below-optimal cross-entropy on the training set.
Test cross-entropy is infinite because the model assigns a categorical zero probability to some possible targets. 
Both L1 and L2 regularizations indirectly constrain the encoding length of the resulting networks and have the advantage of being compatible with backpropagation search. However, these constraints are not as effective as pure MDL in avoiding overfitting (cross-entropy is below optimal on the training set and above on the test set).

\begin{table}[h]
\small
\begin{tabularx}{\columnwidth}{Y||r@{\;}r|r@{\;}r}
\hline
Objective
& \multicolumn{2}{c|}{CE ($\times 10^{-2}$)}
& \multicolumn{2}{c}{Size}
\\ 
function
& \multicolumn{1}{c}{train}
& \multicolumn{1}{c|}{test}
& units
& conn\\ \hline

$|G|$                                                                   & 158.5                                                                               & 158.5                                                                              & 0 & 0                                                                       \\ \hline
$|D:G|$                                                                 & \textbf{37.3}                                                                               & $\infty$                                                                          & 126 & 299                                                                    \\ \hline
$|D:G| + L1$                                                                 & 37.6                                                                              & 55.3                                                                              & 6 & 23                                                                      \\ \hline
$|D:G| + L2$                                                                 & 37.5                                                                               & $\infty$                                                                          & 6 & 33                                                                           \\ \hline
\textbf{$|D:G|+|G|$ (MDL)} & 38.1                                                                      & \textbf{25.8}                                                                     & 1 & 7                                                              \\ \hline
\end{tabularx}

\caption{Cross-entropy and number of units and connections on the $a^nb^n$ task using different objective functions; MDL yields ground-truth optimal CE for both training and test.
}
\label{table:partial_mdl}
\end{table}

\section{Conclusion}

Classical RNNs optimized for accuracy can partially recognize nonregular languages and generalize beyond the data up to a certain $n$ \citep{gers_lstm_2001, weiss:2018}. However large this $n$ may be, the failure of these networks to fully generalize to arbitrary values of $n$ reveals that they fail to lock in on the correct grammars that underlie these tasks.

We found that an MDL-optimized learner arrives at networks that are reliably close to the true distribution with small training corpora, for classically challenging tasks. In several cases, the networks achieved perfect scores. Beyond the usual evaluation in terms of performance on test sets, the networks lent themselves to direct inspection and showed an explicit statement of the pattern that generated the corpus. 

\section*{Acknowledgements}

We wish to thank Matan Abudy, Moysh Bar-Lev, Artyom Barmazel, Marco Baroni, Adi \mbox{Behar-Medrano}, Maxime Caut\'e, Rahma Chaabouni, Emmanuel Dupoux, Nicolas Gu\'erin, Jean-R\'emy King, Yair Lakretz, Tal Linzen, A\"el Quelennec, Ezer Rasin, Mathias Sabl\'e-Meyer, Benjamin Spector; the audiences at CNRS/ENS Paris, Facebook AI Paris, NeuroSpin, Tel Aviv University, and ZAS Berlin; and Nitzan Ron for creating the figures in this paper. We also thank the action editors at TACL and three anonymous reviewers for their helpful comments.

This work was granted access to the HPC/AI resources of IDRIS under the allocation 2021- A0100312378 made by GENCI.

\bibliographystyle{acl_natbib}
\bibliography{mdlnn}

\end{document}